%% file: async_forecast.tex
\documentclass[letterpaper]{article} 
\usepackage{aaai23}[submission]  

\input{macros}

\newcommand{\hnode}[2]{h^{\text{node}, {#1}}_{#2}}
\newcommand{\hedge}[2]{h^{\text{edge}, {#1}}_{#2}}
\newcommand\gnn{\text{gnn}}
\newcommand\node[1][]{^{\text{node}#1}}
\newcommand\edge[1][]{^{\text{edge}#1}}

\floatstyle{ruled}
\newfloat{listing}{tb}{lst}{}
\floatname{listing}{Listing}
\usepackage{url}

\makeatletter
\def\@copyrightspace{}
\makeatother

\title{GraFITi: Graphs for Forecasting Irregularly Sampled Time Series}

\author {
	Vijaya Krishna Yalavarthi,\textsuperscript{\rm 1}
	Kiran Madhusudhanan, \textsuperscript{\rm 1}
	Randolf Sholz \textsuperscript{\rm 1}
	Nourhan Ahmed,\textsuperscript{\rm 1}
	Johannes Burchert, \textsuperscript{\rm 1}
	Shayan Javed \textsuperscript{\rm 1}
	Stefan Born \textsuperscript{\rm 2}
	Lars Schmidt-Thieme \textsuperscript{\rm 1}
}
\affiliations {
	\textsuperscript{\rm 1} ISMLL, University of Hildesheim, Germany\\
	\{yalavarthi, kiranmadhusud, sholz, ahmedn, burchert, shayan,  schmidt-thieme\}@ismll.uni-hildesheim.de,\\
	\textsuperscript{\rm 2} Institute of Mathematics, TU Berlin, Germany\\  born@math.tu-berlin.de
}
\begin{document}\sloppy
	
\maketitle

\input{section-0-absract}
\input{section-1-intro}
\input{section-2-related}

\input{new-section-3-ts-as-graphs}

\input{new-section-5-vijaya_model}

\input{section-6-experiments}
\input{section-7-conclusions}
\bibliography{bib}
\bibliographystyle{aaai24}
\input{apdx}
\end{document}

%% file: macros.tex
\usepackage{dsfont}
\usepackage{mathtools}

\newcommand\R{{\mathds{R}}}

\usepackage{times}  
\usepackage{helvet}  
\usepackage{courier}  
\usepackage[hyphens]{url}  
\usepackage{graphicx} 
\usepackage{natbib}  
\usepackage{caption} 
\DeclareCaptionStyle{ruled}{labelfont=normalfont,labelsep=colon,strut=off} 

\newcommand{\nan}{\texttt{NaN}}
\newcommand{\cupdot}{\mathbin{\dot{\cup}}}

\usepackage{arydshln}
\usepackage[notrig]{physics}
\usepackage{algorithm}
\usepackage[noend]{algorithmic}

\usepackage{newfloat}
\usepackage{listings}
\usepackage{natbib}
\usepackage{tikz}
\usepackage{pgfplots}
\usetikzlibrary{arrows.meta}
\pgfplotsset{compat = newest}
\usepackage{xcolor}
\usetikzlibrary{shapes}
\usepackage{tikz-network}

\makeatletter
\newcommand\footnoteref[1]{\protected@xdef\@thefnmark{\ref{#1}}\@footnotemark}
\makeatother

\usepackage{varwidth}
\usepackage{graphicx}
\usepackage{booktabs}
\usepackage{amsmath}
\usepackage{amsthm}
\usepackage{amsfonts}
\usepackage{epsfig}
\usepackage{graphicx}
\usepackage{subfigure}
\usepackage[bottom]{footmisc}

\usepackage{multirow}
\usepackage{xspace}
\usepackage[singlelinecheck=off]{caption}
\DeclareCaptionType{copyrightbox}
\usepackage{pifont}
\newcommand{\spara}[1]{\smallskip\noindent{\bf #1}}

\newcommand{\squishlist}{
 \begin{list}{$\bullet$}
  {  \setlength{\itemsep}{0pt}
     \setlength{\parsep}{3pt}
     \setlength{\topsep}{3pt}
     \setlength{\partopsep}{0pt}
     \setlength{\leftmargin}{2em}
     \setlength{\labelwidth}{1.5em}
     \setlength{\labelsep}{0.5em}
} }
\newcommand{\squishlisttight}{
 \begin{list}{$\bullet$}
  { \setlength{\itemsep}{0pt}
    \setlength{\parsep}{0pt}
    \setlength{\topsep}{0pt}
    \setlength{\partopsep}{0pt}
    \setlength{\leftmargin}{2em}
    \setlength{\labelwidth}{1.5em}
    \setlength{\labelsep}{0.5em}
} }

\newcommand{\squishdesc}{
 \begin{list}{}
  {  \setlength{\itemsep}{0pt}
     \setlength{\parsep}{3pt}
     \setlength{\topsep}{3pt}
     \setlength{\partopsep}{0pt}
     \setlength{\leftmargin}{1em}
     \setlength{\labelwidth}{1.5em}
     \setlength{\labelsep}{0.5em}
} }

\newcommand{\squishend}{
  \end{list}
}

\newcommand{\eat}[1]{}

\newcounter{ccc}

\newcommand{\bigO}{\mathcal{O}}

%% file: section-0-absract.tex
\begin{abstract}
	Forecasting irregularly sampled time series with missing values
	is a crucial task for numerous real-world applications such as
	healthcare, astronomy, and climate sciences.
	State-of-the-art approaches to this problem rely on Ordinary Differential Equations (ODEs)
	which are known to be slow and often require additional features to handle missing values.
	To address this issue, we propose a novel model
	using \underline{Gra}phs for \underline{F}orecasting \underline{I}rregularly Sampled \underline{Ti}me Series with missing values
	which we call GraFITi.
	GraFITi first converts the time series to a Sparsity Structure Graph
	which is a sparse bipartite graph,
	and then reformulates the forecasting problem as the edge weight prediction task in the graph.
	It uses the power of Graph Neural Networks to learn the graph
	and predict the target edge weights.
	GraFITi has been tested on $3$ real-world and $1$ synthetic irregularly sampled time series dataset with missing values
	and compared with various state-of-the-art models.
	The experimental results demonstrate that GraFITi improves the forecasting accuracy by up to $17\%$
	and reduces the run time up to $5$ times compared to the state-of-the-art forecasting models.
\end{abstract}

%% file: section-1-intro.tex
\section{Introduction}
\label{sec:intro}

\begin{figure}[t]
	\subfigure[Forecasting regular multivariate time series (MTS)]{
		\centering
		\includegraphics[width=0.8\linewidth, clip]{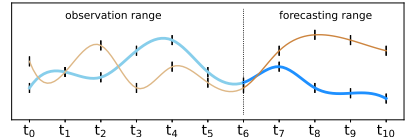}
	}
	\subfigure[Forecasting irregular multivariate time series]{
		\centering
		\includegraphics[width=0.8\columnwidth, clip]{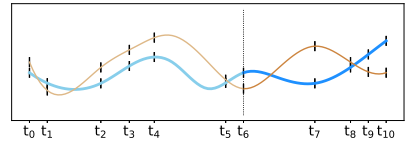}
	}
	\subfigure[Forecasting irregularly sampled multivariate time series with missing values (IMTS)]{
		\centering
		\includegraphics[width=0.8\columnwidth, clip]{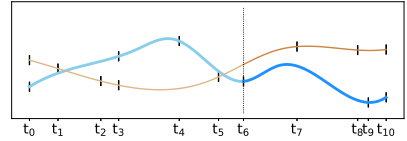}
	}
	\caption{(a) multivariate time series forecasting, (b) irregular multivariate time series forecasting, (c) forecasting irregularly sampled multivariate time series with missing values. In all cases the observation range is from time $t_0$ to $t_6$ and the forecasting range is from time $t_7$ to $t_{10}$.}
	\label{fig:intro}
\end{figure}

Time series forecasting predicts future values based on past observations.
While extensively studied, 
most research focuses on regularly sampled and fully observed multivariate time series (MTS)~\cite{LZ21,ZC22,DJ06}.
Limited attention is given to irregularly sampled time series with missing values (IMTS) which is commonly seen in many real-world applications.
IMTS has independently observed channels at irregular intervals, resulting in sparse data alignment.
The focus of this work is on forecasting IMTS. Additionally, there is another type called irregular multivariate time series which is
fully observed but with irregular time intervals (Figure~\ref{fig:intro} illustrates the differences)
which is not the interest of this paper.

Ordinary Differential Equations (ODE) model continuous time series predicting system evolution over time based on the rate of change of state variables as shown in Eq.~\ref{eq:ode}. 
\begin{align}
	\frac{\mathrm{d}}{\mathrm{d}t}x(t) = f(t,x(t)) \label{eq:ode}
\end{align}
ODE-based models~\cite{SE22,DS19,BS21,A23} are able to forecast at arbitrary time points.
However, ODE models can be slow 
because of their auto-regressive nature and computationally expensive numerical integration process. 
Also, some ODE models cannot directly handle missing values in the observation space, hence, they often rely on missing value indicators~\cite{DS19,BS21} which are given as additional channels in the data.

In this work, we propose a novel model called GraFITi: graphs for forecasting IMTS. GraFITi converts IMTS data into a Sparsity Structure Graph and formulates forecasting as edge weight prediction in the graph. This approach represents channels and timepoints as disjoint nodes connected by edges in a bipartite graph. GraFITi uses multiple graph neural network (GNN) layers with attention and feed-forward mechanisms to learn node and edge interactions. Our Sparsity Structure Graph, by design, provides a more dynamic and adaptive approach to process IMTS data, and improves the performance of the forecasting task.

We evaluated GraFITi for forecasting IMTS using $3$ real-world and $1$ synthetic dataset.
Comparing it with state-of-the-art methods for IMTS and selected baselines for MTS,
GraFITi provides superior forecasts.

Our contributions are summarized as follows:
\begin{itemize}
	
	\item We introduce a novel representation of irregularly sampled multivariate time series with
	 missing values (IMTS) as sparse bipartite graphs, the Sparsity Structure Graph, that efficiently
	  can handle missing values in the observation space of the time series (Section~\ref{sec:ssgraph}).
		
	\item We propose a novel model based on this representation, GraFITi, that can leverage any graph
	 neural network to perform time series forecasting for IMTS (section~\ref{sec:proposed}).
	
	\item We provide extensive experimental evaluation on $3$ real world and $1$ synthetic dataset that
	 shows that GraFITi improves the forecasting accuracy of the best existing models by up to $17\%$
	  and the run time improvement up to $5$ times (section~\ref{sec:exp}).
\end{itemize}
We provide the implementation at {\url{https://anonymous.4open.science/r/GraFITi-8F7B}}.

%% file: section-2-related.tex
\section{Related Work}
\label{sec:related}

This work focus on the forecasting of irregularly sampled multivariate time series data with missing values using graphs. In this section, we discuss the research done in: forecasting models for IMTS, Graphs for forecasting MTS, and models for edge weight prediction in graphs.

\paragraph{Forecasting of IMTS}

Research on IMTS has mainly focused on classification~\cite{LM15,LK16,RC19,SM21,HM20,TS21} and interpolation~\cite{CP18,RC19,SM21,TS21,SM22,YB22}, with limited attention to forecasting tasks. Existing models for these tasks mostly rely on Neural ODEs~\cite{CP18}. In Latent-ODE~\cite{RC19}, an ODE was combined with a Recurrent Neural Network (RNN) for updating the state at the point of new observation. The GRU-ODE-Bayes model~\cite{DS19} improved upon this approach by incorporating GRUs, ODEs, and Bayesian inference for parameter estimation. The Continuous Recurrent Unit (CRU)~\cite{SE22} based model uses a state-space model with stochastic differential equations and kalman filtering. The recent LinODENet model~\cite{A23} enhanced CRU by using linear ODEs and ensure self-consistency in the forecasts. Another branch of study involves Neural Flows~\cite{BS21}, which use neural networks to model ODE solution curves, rendering the ODE integrator unnecessary. Among various flow architectures, GRU flows have shown good performance.

\paragraph{Using graphs for MTS}

In addition to CNNs, RNNs, and Transformers, graph-based methods have been studied for IMTS forecasting.
Early GNN-based approaches, such as \cite{WP20},
required a pre-defined adjacency matrix to establish relationships between the time series channels.
More recent models like the Spectral Temporal Graph Neural Network~\cite{CW20} (STGNN) and
the Time-Aware Zigzag Network~\cite{CS21} improved on this
by using GNNs to capture dependencies between variables in the time series.
On the other hand, \citet{SR22} proposed a bipartite setup with induced nodes to reduce graph complexity, built solely from the channels.
Existing graph-based time series forecasting models focus on learning correlations or similarities between channels,
without fully exploiting the graph structure.
Recently, GNNs were used for imputation and reconstruction of MTS with missing values,
treating MTS as sequences of graphs where nodes represent sensors and edges denote correlation~\cite{AI22,TA22}.
Similar to previous studies, they learn similarity or correlation among channels.

\paragraph{Graph Neural Networks for edge weight prediction}

Graph Neural Networks (GNNs) are designed to process graph-based data. While most GNN literature such as Graph Convolutional Networks, Graph Attention Networks focuses on node classification~\cite{KN17,VC17}, a few studies have addressed edge weight prediction. Existing methods ~\cite{DP11,FZ18} in this domain rely on latent features and graph heuristics, such as node similarity~\cite{ZM15}, proximity measures~\cite{MM07}, and local rankings~\cite{XB20}. Recently, deep learning-based approaches~\cite{HH17,ZS22,YM20} were proposed. Another branch of research deals with edge weight prediction in weighted signed graphs~\cite{KS16} tailored to social networks. However, all proposed methods typically operate in a transductive setup with a single graph split into training and testing data, which may not be suitable for cases involving multiple graphs like ours, where training and evaluation need to be done on separate graph partitions.

\begin{figure}[ht]
	\centering
	\includegraphics[width=0.9\linewidth, clip]{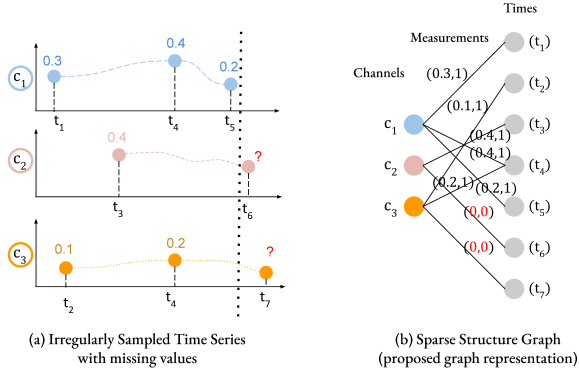}
	\caption{Representation of IMTS as Sparsity Structure Graph. (b) is the Sparsity Structure Graph representation of (a) where times and channels are the nodes and observation measurements are the edges with observations values. Target edges are provided with $\color{red} (0,0)$.}
	\label{fig:graph_rep}
\end{figure}

%% file: new-section-3-ts-as-graphs.tex
\section{The Time Series Forecasting Problem}
An \textbf{irregularly sampled multivariate times series with missing values},
is a finite sequence of pairs $S = (t_n, x_n)_{n=1:N}$ where
$t_n\in\R$ is the $n$-th \textbf{observation timepoint} and
$x_n \in (\R\cup\{\nan\})^C$ is the $n$-th \textbf{observation event}. Components with
$x_{n,c} \neq \nan$ represent \textbf{observed values} by channel $c$ at event time $t_n$, and
$x_{n,c}= \nan$ represents a missing value. $C$ is the total number of channels.

A \textbf{time series query} is a pair $(Q, S)$ of
a time series $S$ and a sequence $Q=(q_k, c_k)_{k=1:K}$ such that the value of channel $c_k\in\{1,\ldots,C\}$ is to be predicted at time $q_k\in\R$.
We call a query a \textbf{forecasting query}, if all its
query timepoints are after the last timepoint of the time
series $S$, an \textbf{imputation query} if all of them are
before the last timepoint of $S$ and a \textbf{mixed query}
otherwise. In this paper, we are interested in forecasting only.

A vector $y\in\R^K$ we call an \textbf{answer} to the forecasting query:
$y_k$ is understood as the predicated value of time series $S$ at time
$q_k$  in channel $c_k$.
The difference between two answers $y, y'$ to the same query can be measured by any loss function, for example by a simple squared error
\begin{align*}
	\ell(y, y') \coloneqq  \frac{1}{K} \sum_{k=1}^K (y_k - y_k')^2
\end{align*}

The \textbf{time series forecasting problem} is as follows:
given
a dataset of pairs $D\coloneqq (Q_i, S_i, y_i)_{i=1:M}$ of forecasting queries and ground truth answers
from an unknown distribution $p^\text{data}$ and
a loss function $\ell$ on forecasting answers,
find a forecasting model $\hat y$
that maps queries $(Q, S)$ to answers $\hat{y}(Q, S)$
such that
the expected loss between ground truth answers and forecasted
answers is minimal:
\begin{align*}
	\mathcal{L}(\hat y; p^\text{data}) \coloneqq  \mathbb{E}_{(Q,S,y)\sim p^\text{data}} \big[\ell(y, \hat y(Q, S))\big]
\end{align*}

\section{Sparsity Structure Graph Representation}
\label{sec:ssgraph}

We describe the proposed Sparsity Structure Graph representation and convert the forecasting problem as edge weight prediction problem. Using the proposed representation:

\begin{itemize}
	\item We explicitly obtain the relationship between the channels and times via observation values allowing the inductive bias of the data to pass into the model.
	\item We elegantly handle the missing values in IMTS in the observation space by connecting edges only for the observed values.
\end{itemize}

Missing values represented by \nan-values are unsuited for standard arithmetical operations. Therefore, they are often encoded by dedicated binary variables called \textbf{missing value indicators} or masks:
$x_n \in (\R \times \{0,1\})^C$.
Here, $(x_{n,c,1},1)$ encodes an observed value and $(0, 0)$ encodes a missing value.
Usually, both components are seen as different scalar variables:
the real value $x_{n,c,1}$ and its binary missing value indicator / mask $x_{n,c,2}$,
the relation between both is dropped and observations simply 
modeled as $x_n\in\R^{2C}$.

We propose a novel representation of a time series $S$ using a bipartite graph $G = (V, E)$.
The graph has nodes for channels and timepoints, denoted as $V_C$ and $V_T$ respectively ($V = V_C \cupdot V_T$). Edges $E \subseteq V_C\times V_T$ in the graph connect each channel node to its corresponding timepoint node with an observation. Edge features $F\edge$ are the observation values and node features $F\node$ are the channel IDs and timepoints.
Nodes $V_C \coloneqq \{1,\ldots, C\}$ represent channels and nodes $V_T\coloneqq \{C+1, \ldots, C+N\}$ represent unique timepoints:
\begin{align}%
	\small
	\label{eq: label}%
	\begin{aligned}
		V  & \coloneqq  \{1,\ldots, C+N\}=V_C \cupdot V_T
		\\ E  & \coloneqq  \bigl\{\{i, j\} \mid x_{i-C,j}\neq \nan, i\in V_T, j\in V_C\bigr\}
		\\ F^{\text{node}}_v & \coloneqq  \begin{cases}
			v       \hfill:& v\in V_C
			\\ t_j     \hfill:& v\in V_T, \;j=v-C
		\end{cases}
		\\ F^{\text{edge}}_e & \coloneqq  x_{i-C,j} \qq{for} e = \{i, j\}\in E \qq{with} i\in V_T, j\in V_C
	\end{aligned}
\end{align}
For an IMTS, missing values make the bipartite graph sparse, meaning $|E| \ll C\cdot N$. However, for a fully observed time series, where there are no missing values, i.e. $|E| = C\cdot N$, the graph is a complete bipartite graph.

We extend this representation to time series queries $(S,Q)$
by adding additional edges between queried channels and timepoints,
and distinguish observed and queried edges by an additional binary edge feature called target indicator. 
{\em Note that the target indicator
used to differentiate the observed edge and target edge
is different from the missing value indicator
which is used to represent the missing observations in the observation space.}
Given a query $Q = (q_k, c_k)_{k=1:K}$, let $(t'_1,\ldots,t'_{K'})$ be an enumeration of the unique queried timepoints $q_k$.
We introduce additional nodes $V_Q\coloneqq \{C+N+1, \ldots, C+N+K'\}$ so that the augmented graph, together with the node and edge features is given as
\begin{align}
	\begin{aligned}
		V  & \coloneqq   V_C\cupdot V_T \cupdot V_Q = \{1,\ldots, C+N+K'\}
		\\  E  & \coloneqq  \bigl\{   \{i, j\} \mid x_{i-C,j}\neq \nan,  i\in V_{T}, j\in V_C  \bigr\}
		\\        & \quad\cup \bigl\{ \{ i, j \} \mid   i\in V_{Q}, j\in V_C, (t'_{i-N-C},j) \in Q  \bigr\}
		\\ F^{\text{node}}_v & \coloneqq  \begin{cases}
			v   \hfill:&  v\in V_C
			\\ t_j \hfill:&  v\in V_{T}, j = v-C
			\\ t'_j \hfill:&  v\in V_{Q}, j = v-C-N
		\end{cases}
		\\ F^{\text{edge}}_e&\coloneqq
		\begin{cases}
			(x_{i,j,1}, 1) \hfill:& e=\{i,j\}\in E , \quad i\in V_T, j\in V_C  \\
			(0, 0)       \hfill:& e=\{i,j\}\in E , \quad i\in V_Q, j\in V_C  \\
		\end{cases}
	\end{aligned}  \;
\end{align}
where $(t'_{i-N-C},j) \in Q$ is supposed to mean that $(t'_{i-N-C},j)$ appears in the sequence $Q$.
To denote this graph representation, we write briefly
\begin{align}
	\text{ts2graph}(X,Q) \coloneqq  (V,E,F^{\text{node}},F^{\text{edge}} ) \label{eq:ts2graph}
\end{align}
The conversion of an IMTS to a Sparsity Structure Graph is shown in Figure~\ref{fig:graph_rep}.

To make the graph representation $(V,E,F\node,F\edge)$
of a time series query
processable by a graph neural network, node and edge features have to be
properly embedded, otherwise, both, the nominal channel ID and the timepoint are
hard to compute on.  We propose an \textbf{Initial Embedding layer}
that encodes
channel IDs via a onehot encoding and
time points via a learned sinusoidal encoding~\cite{SM21}:
\begin{align}
	h_v\node[,0] & \coloneqq
	\begin{cases}
		\mathbf{FF}(\text{onehot}(F\node_v)) &: v\in V_C
		\\ \sin(\mathbf{FF}(F\node_v))          &: v\in V_T\cupdot V_Q
	\end{cases} \label{eq:init_node}
	\\
	h_e\edge[,0] & \coloneqq  \mathbf{FF}(F\edge_e) \qq{for} e\in E \label{eq:init_edge}
\end{align}
where onehot denotes the binary indicator vector and
$\mathbf{FF}$ denotes a separate fully connected layer in each case.

The final graph neural network layer $(h\node[,L], h\edge[,L])$
has embedding dimension $1$. The scalar values of the query edges are taken as the
predicted answers to the encoded forecasting query:
\begin{align}\label{eq:graph2ts}
	\hat y & \coloneqq  \text{graph2ts}(h\node[,L], h\edge[,L], V, E) = (h^{\text{edge},L}_{e_k})_{k=1:K} \notag
	\\ &\phantom{{}\coloneqq{}} \text{where } e_k=\{C+N+k', c_k\} \qq{with} t'_{k'}=q_k 
\end{align}

%% file: new-section-5-vijaya_model.tex
\section{Forecasting with GraFITi}
\label{sec:proposed}


GraFITi first encodes the time series query to graph using Eq.~\ref{eq:ts2graph} and compute initial embeddings for the nodes ($\hnode{0}{}$) and edges ($\hedge{0}{}$) using Eqs.~\ref{eq:init_node} and~\ref{eq:init_edge} respectively.
Now, we can leverage the power of graph neural networks for
further processing the encoded graph. Node and edge features
are updated layer wise, from layer $l$ to $l+1$ using a graph neural network:
\begin{align}
	(\hnode{l+1}{}, \hedge{l+1}{}) \coloneqq
	\gnn^{(l)}(\hnode{l}{}, \hedge{l}{}, V,E) \label{eq:gnn(l)}
\end{align}

\begin{figure*}
	\centering
	\includegraphics[width=\linewidth]{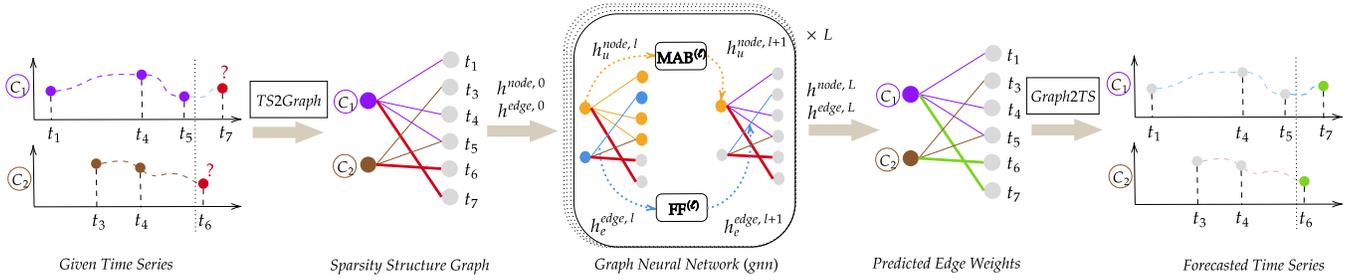}
	\caption{Overall architecture of GraFITi.}
	\label{fig:architecture}
\end{figure*}

There have been a variety of {\gnn} architectures such as Graph Convolutional Networks~\cite{KN17},
Graph Attention Networks~\cite{VC17}, proposed in the literature.
In this work, we propose a model adapting the Graph Attention Network~\cite{VC17} to our graph setting
and incorporate essential components for handling sparsity structure graphs.
While a Graph Attention Network computes attention weights by adding queries and keys,
we found no advantage in using this approach (see supplementary material).
Thus, we utilize standard attention mechanism, in our attention block,
as it has been widely used in the literature~\cite{ZZ21}.
Additionally, we also use edge embeddings in our setup to update node embeddings in a principled manner.

\spara{Graph Neural Network (\gnn)}
\label{sec:mhnetwork}

First, we define Multi-head Attention block (\textbf{MAB}) and Neighborhood functions that are used in our {\gnn}.

A {Multi-head attention block} (\textbf{MAB})~\cite{VS17} is represented as:
\begin{align}
	\textbf{MAB}(\mathcal{Q}, \mathcal{K}, \mathcal{V}) &\coloneqq  \alpha(\mathcal{H} + \textbf{FF}(\mathcal{H})) \nonumber\\
	\text{where} \quad  \mathcal{H} &\coloneqq  \alpha(\mathcal{Q} + \textbf{MHA}(\mathcal{Q}, \mathcal{K}, \mathcal{V})) \label{eq:MAB}
\end{align}
where $\mathcal{Q}, \mathcal{K}$ and, $\mathcal{V}$ are called queries, keys, and values respectively, \textbf{MHA} is multi-head attention~\cite{VS17}, $\alpha$ is a non-linear activation.

The Neighborhood of a node $u$ is defined as the set of all the nodes connected to $u$ through edges in $E$:
\begin{align}
	\mathcal{N}(u) \coloneqq  \{v \mid \{u,v\} \in {E}\}
\end{align}

GraFITi consists of $L$ \gnn layers. In each layer, node embeddings are updated using neighbor node embeddings and edge embeddings connecting them. For edge embeddings, we use the embeddings of the adjacent nodes and the current edge embedding. The overall architecture of GraFITi is shown in Figure~\ref{fig:architecture}.

\paragraph{Update node embeddings}
To update embedding of a node $u \in V$, first, we create a sequence of features $H_u$ concatenating its neighbor node embedding $\hnode{l}{v}$ and edge embedding $\hedge{l}{e}, e = \{u,v\}$ where $v \in \mathcal{N}(u)$. We then pass $\hnode{l}{u}$ as queries and $H_u$ as keys and values to $\mathbf{MAB}$.
\begin{align}
\hnode{l+1}{u} & \coloneqq  \textbf{MAB}^{(l)}\left(\hnode{l}{u}, H_u, H_u\right) \label{eq:node_embed}
\\ H_u & \coloneqq \bigl( [\hnode{l}{v} \mathbin\Vert \, \hedge{l}{e}]\bigr)_{v \in \mathcal{N}(u)} ,\; e=\{u,v\}
\end{align}

\paragraph{Updating edge embeddings:}
To compute edge embedding $\hedge{l+1}{e}$, $e = \{u,v\}$ we concatenate $\hnode{l}{u}, \hnode{l}{v}$ and $\hedge{l}{e}$, and  pass it through a dense layer ($\mathbf{FF}$) followed by a residual connection and nonlinear activation.
\begin{align}
\hedge{l+1}{e} \coloneqq  \alpha\left(\hedge{l}{e} + \mathbf{FF}^{(l)}\left(\hnode{l}{u}\mathbin\Vert \hnode{l}{v} \mathbin\Vert \hedge{l}{e}\right)\right) \label{eq:edge_embed}
\end{align}
where $e = \{u,v\}$.
{\em Note that, although our edges are undirected, we compute the edge embedding by concatenating the embeddings in a specific order i.e., the channel embedding, time embedding and edge embedding.}
We show the process of updating nodes and edges in layer $l$ using a \gnn in Algorithm~\ref{alg:gnn}.

\begin{algorithm}[t]
	\caption{Graph Neural Network ($\gnn^{(l)}$)}
	\small
	\label{alg:gnn}
	\begin{algorithmic}
		\REQUIRE $\hnode{l}{}, \hedge{l}{}, V, E$
		\FOR {$u \in {V}$}
		\STATE $H_u \gets  \bigl( [\hnode{l}{v} \mathbin\Vert \, \hedge{l}{e}]\bigr)_{v \in \mathcal{N}(u)}$ \hfill \COMMENT{$e=\{u,v\}$}
		\STATE $\hnode{l+1}{u} \gets  \textbf{MAB}^{(l)}(\hnode{l}{u}, H_u, H_u)$
		\ENDFOR
		
		\FOR {$e=\{u,v\} \in {E}$}
		\STATE $\hedge{l+1}{e} \gets  \alpha \left(\hedge{l}{e} + \mathbf{FF}^{(l)}\left(\bigl[\hnode{l}{u} \mathbin\Vert \hnode{l}{v} \mathbin\Vert \hedge{l}{e}\bigr]\right) \right)$
		\ENDFOR
		
		\RETURN {$\hnode{l+1}{}, \hedge{l+1}{}$}
	\end{algorithmic}
\end{algorithm}

\paragraph{Answering the queries:}
\label{sec:target_pred}
As mentioned in Section~\ref{sec:ssgraph}, our last ${\gnn}^{(L)}$ layer has embedding dimension $1$. Hence,
after processing the graph features through $L$ many $\gnn$ layers, we use Eq.~\ref{eq:graph2ts} to decode the graph and provide the predicted answers to the time series query. A forward pass of GraFITi is presented in Algorithm~\ref{alg:gratif}.
\begin{algorithm}[t]
	\caption{Forward pass of GraFITi}
	\small
	\label{alg:gratif}
	\begin{algorithmic}
		\REQUIRE Observed time series forecasting query $(S,Q)$

		\STATE $(V,E,F_{\text{node}}, F_{\text{edge}}) \gets \text{ts2graph}(S,Q)$ \hfill \COMMENT{Eq.~\ref{eq:ts2graph}}

		\COMMENT{{\textbf{Initial embeddings of nodes and edges}}}

		\STATE $\hnode{0}{} \gets  \{\hnode{0}{u} \mid u \in {V} \}$ \hfill \COMMENT{Eq.~\ref{eq:init_node}}

		\STATE $\hedge{0}{} \gets  \{\hedge{0}{u,v} \mid \{u,v\} \in {E}\}$ \hfill	\COMMENT{Eq:~\ref{eq:init_edge}}

		\COMMENT{\textbf{Graph Neural Network}}

		\FOR {$l \in \{1, \ldots ,L\}$}

		\STATE $\hnode{l+1}{}, \hedge{l+1}{} \gets \gnn^{(l)}(\hnode{l}{}, \hedge{l}{}, V, E)$  \hfill \COMMENT{Alg.~\ref{alg:gnn}}

		\ENDFOR

		\STATE $\hat{y} \gets \text{graph2ts}(\hnode{L}{}, \hedge{L}{}, V, E)$ \hfill \COMMENT{Eq:~\ref{eq:graph2ts}}

		\RETURN $\hat{y}$

	\end{algorithmic}
\end{algorithm}

\paragraph{Computational Complexity:}
\label{sec:complexity}

The computational complexity of GraFITi primarily comes from using MAB in Eq.~\ref{eq:node_embed}.
For a single channel node $u \in \{1,...,C\}$,
the maximum complexity for computing its embedding is $\mathcal{N}(u)$
since only neighborhood connections are used for the update, and $\mathcal{N}(u) \subseteq \{C+1,...,C+N+K'\}$.
Thus, computing the embeddings of all channel nodes is $\bigO(|E|)$.
Similarly, the computational complexity of MAB for computing the embeddings of all nodes in $V_T \cupdot V_Q$ is also $\bigO(|E|)$.
A feed forward layer $\mathbf{FF}:\R^Y \to \R^Z$ will have a computational complexity of $\bigO(YZ)$.


\paragraph{Delineating from GRAPE~\cite{YM20}} \citet{YM20} introduced GRAPE,
a graph-based model for imputing and classifying vector datasets with missing values.
This approach employs a bipartite graph, with nodes divided into separate sets for sample IDs and sample features.
The edges of this graph represent the feature values associated with the samples.
Notably, GRAPE learns in a transductive manner, encompassing all the data samples,
including those from the test set, within in the graph.
In contrast, GraFITi uses inductive approach.
Here, each instance is a Sparsity Structure Graph, tailored for time series data.
In this structure, nodes are divided into distinct sets for channels and timepoints,
while the edges are the time series observations.

%


%% file: section-6-experiments.tex
\section{Experiments}
\label{sec:exp}


\begin{table}[ht]
	\centering
	\scriptsize
	\caption{Statistics of the datasets used in the experiments. Sparsity means the percentage of missing observations in the time series}
	\label{tab:dset}
	\begin{tabular}{lccccc}
		\hline
		Name & \#Sample & \#Chann. & Max.len. & Max.Obs.& Sparsity \\
		\hline
		USHCN & 1,100 & 5 & 290& 320 &$77.9\%$\\
		MIMIC-III & 21,000& 96 & 96 & 710 &$94.2\%$\\
		MIMIC-IV & 18,000 & 102 & 710 & 1340 &$97.8\%$\\
		Physionet'12 & 12,000 & 37 & 48 & 520 &$85.7\%$\\
		\hline
	\end{tabular}
\end{table}

\subsection{Dataset description}
\label{dset}

$4$ datasets including $3$ real world medical and $1$ synthetic climate IMTS datasets are used for evaluating the proposed model.
Basic statistics of the datasets is provided in Table~\ref{tab:dset}.

{Physionet'12}~\cite{SM12} consists of ICU patient records observed for $48$ hours. 
{MIMIC-III}~\cite{JP16} is also a medical dataset that consists measurements of the ICU patients observed for $48$ hours.
{MIMIC-IV}~\cite{JB21} is built upon the MIMIC-III database.
{USHCN}~\cite{MW15} is a climate dataset that consists of the measurements of
daily temperatures, precipitation and snow observed over $150$ years from $1218$ meteorological stations in the USA.
For MIMIC-III, MIMIC-IV and USHCN, we followed the pre-processing steps provided by~\citet{A23,BS21,DS19}.
Hence, observations in MIMIC-III and MIMIC-IV are rounded for $30$ mins and $1$ hour respectively.
Whereas for the Physionet'12,
we follow the protocol of~\citet{CP18,CW18,TS21} and processed the dataset to have hourly observations. 

\begin{table*}[t]
	\centering
	\scriptsize
	\caption{Experimental results for forecasting next three time steps. Evaluation metric MSE, Lower is better. Best results are in {bold} and the next best are in {italics}. Published results are presented in open brackets, Physionet'12 dataset was not used by the baseline models hence do not have published results. We show $\%$ improvement with $\uparrow$. `\textnormal{ME}' indicates Memory Error.}
	\label{tab:base_exp}
	\begin{tabular}{l|l|l|l|l}
		\hline  & \multicolumn{1}{|c}{USHCN} & \multicolumn{1}{|c}{MIMIC-III} & \multicolumn{1}{|c|}{MIMIC-IV} & Physionet'12\\ 
		\hline
		DLinear+ & $0.347 \pm 0.065$ & $0.691 \pm 0.016$ & $0.577 \pm 0.001$ &$0.380 \pm 0.001$ \\
		NLinear+ & $0.452 \pm 0.101$ & $0.726 \pm 0.019$ & $0.620 \pm 0.002$ & $0.382 \pm 0.001$  \\
		Informer+ & $0.320 \pm 0.047$ & $0.512 \pm 0.064$ & $0.420 \pm 0.007$&  $0.347 \pm 0.001$\\
		FedFormer+ & $2.990 \pm 0.476$ & $1.100 \pm 0.059$ & $2.135 \pm 0.304$ & $0.455 \pm 0.004$\\
		
		\hdashline
		NeuralODE-VAE & $\quad \quad \; \, -$ \hfill ($0.960 \pm 0.110$) & $\quad \quad \; \, -$ \hfill ($0.890 \pm 0.010$) &  $\quad \quad \; \, -$ & $-$\\
		Sequential VAE & $\quad \quad \; \, -$ \hfill ($0.830 \pm 0.070$) & $\quad \quad \; \, -$ \hfill ($0.920 \pm 0.090$) & $\quad \quad \; \, -$ &$-$\\
		GRU-Simple & $\quad \quad \; \, -$ \hfill ($0.750 \pm 0.120$) & $\quad \quad \; \, -$ \hfill ($0.820 \pm 0.050$) & $\quad \quad \; \, -$ &$-$\\
		GRU-D & $\quad \quad \; \, -$ \hfill ($0.530 \pm 0.060$) & $\quad \quad \; \, -$ \hfill ($0.790 \pm 0.060$) & $\quad \quad \; \, -$ &$-$\\
		T-LSTM & $\quad \quad \; \, -$ \hfill ($0.590 \pm 0.110$) & $\quad \quad \; \, -$ \hfill ($0.620 \pm 0.050$) & $\quad \quad \; \, -$ &$-$\\
		mTAN & $0.300 \pm 0.038$ & $0.540 \pm 0.036$ & $\quad \quad \; $ME & $0.315 \pm 0.002$ \\
		GRU-ODE-Bayes & $0.401 \pm 0.089$ (${0.430 \pm 0.070})$ & $0.476 \pm 0.043$ \hfill (${0.480 \pm 0.010}$) & $0.360 \pm 0.001$ $(0.379 \pm 0.005)$ & $0.329 \pm 0.004$\\
		Neural Flow & $0.414 \pm 0.102$ & $0.477 \pm 0.041$ \hfill (${0.490} \pm 0.004$) & $ 0.354 \pm 0.001$ (${0.364} \pm 0.008$) & $0.326 \pm 0.004$\\
		CRU & $0.290 \pm 0.060$ & $0.592 \pm 0.049$ & $\quad \quad \; $ME & $0.379\pm 0.003$ \\
		LinODEnet & $0.300 \pm 0.060$ ($0.290 \pm 0.060$) & $0.446 \pm 0.033$ \hfill (${0 . 4 50} \pm {0 . 0 20}$) & $0.272 \pm 0.002$ ($0.274 \pm 0.002$) &  $0.299 \pm 0.001$\\
		\hdashline
		GraFITi (ours) & $\mathbf{0.272 \pm 0.047}$ $\quad \mathbf{\uparrow \; 9.3}\%$ & $\mathbf{0.396 \pm 0.030}$ $\quad \mathbf{\uparrow \; 11.2}\%$& $\mathbf{0.225 \pm 0.001}$ $\quad \mathbf{\uparrow \; 17.2\%}$ & $\mathbf{0.286 \pm 0.001}$ $\; \mathbf{\uparrow4.3\%}$\\
	\end{tabular}
\end{table*}

\subsection{Competing algorithms}
\label{sec:baselines}

Here, we provide brief details of the models that are compared with the proposed GraFITi for evaluation.

We select $4$ IMTS forecasting models for comparison, including {GRU-ODE-Bayes}~\cite{DS19}, Neural Flows~\cite{BS21}, CRU~\cite{SE22} and LinODENet~\cite{A23}. Additionally, we use the well established IMTS interpolation model mTAN~\cite{SM21}. It is interesting to verify the performance of well established MTS forecasting models for IMTS setup. We do this by adding missing value indicators as the separate channels to the series and process the time series along with the missing value indicators. Hence we compare with Informer+, Fedformer+, DLinear+ and NLinear+ which are variants of Informer~\cite{ZZ21}, FedFormer~\cite{ZM22}, DLinear and NLinear~\cite{ZC22} respectively. We also compare with the published results from~\cite{DS19} for the NeuralODE-VAE~\cite{CR18}, Sequential VAE~\cite{KS15,KS17}, GRU-Simple~\cite{CP18}, GRU-D~\cite{CP18} and T-LSTM~\cite{BX17}.

\subsection{Experimental setup}
\label{sec:exp_protocol}

\paragraph{Task protocol}

We followed~\citet{A23,BS21,DS19}, applied 5-fold cross-validation and selected hyperparameters using a holdout validation set ($20\%$). For evaluation, we used $10\%$ unseen data. All models were trained on Mean Squared Error, which is also the evaluation metric.

\paragraph{Hyperparamter search}

We searched the following hyperparameters for GraFITi: $L \in \{1,2,3,4\}$, \#heads in \textbf{MAB} from $\{1,2,4\}$, and hidden nodes in dense layers from $\{16,32,64,128,256\}$. We followed the procedure of~\citet{HM20} for selecting the hyperparameters. Specifically, we randomly sampled sets of $5$ different hyperparameters and choose the one that has the best performance on validation dataset. We used the Adam optimizer with learning rate of $0.001$, halving it when validation loss did not improve for 10 epochs. All models were trained for up to $200$ epochs, using early stopping with a patience to $30$ epochs. Hyperparameters for the baseline models are presented in the supplementary material. All the models were experimented using the PyTorch library on a GeForce RTX-3090 GPU.

\subsection{Experimental results}
\label{sec:results}

First, we set the observation and prediction range of the IMTS following~\cite{A23,BS21,DS19}. For the USHCN dataset, the model observes for the first $3$ years and forecasts the next $3$ time steps. For the medical datasets, the model observes for the first $36$ hours in the series and predicts the next $3$ time steps. The results, including the mean and standard deviation, are presented in Table~\ref{tab:base_exp}. The best result is highlighted in bold and the next best in italics. Additionally, we also provide the published results from~\cite{A23,BS21,DS19} in brackets for comparison.

The proposed GraFITi model is shown to be superior compared to all baseline models across all the datasets.
Specifically, in the MIMIC-III and MIMIC-IV datasets,
{\em GraFITi provides around $11.2\%$ and $17.2\%$ improvement
in forecasting accuracy compared to the next best IMTS forecasting model LinODEnet}.
The results on the USHCN dataset have high variance, therefore, it is challenging to compare the models on this dataset.
However, we experimented on it for completeness.
Again, we achieve the best result with $9.2\%$ improvement compared to the next best model.
We note that, the MTS forecasting models that are adapted for the IMTS task,
perform worse than any of the IMTS forecasting models demonstrating the limitation of MTS models applied to IMTS tasks.

\begin{figure}[t]
	\centering
	\subfigure[USHCN]{
		\includegraphics[width=0.45\linewidth]{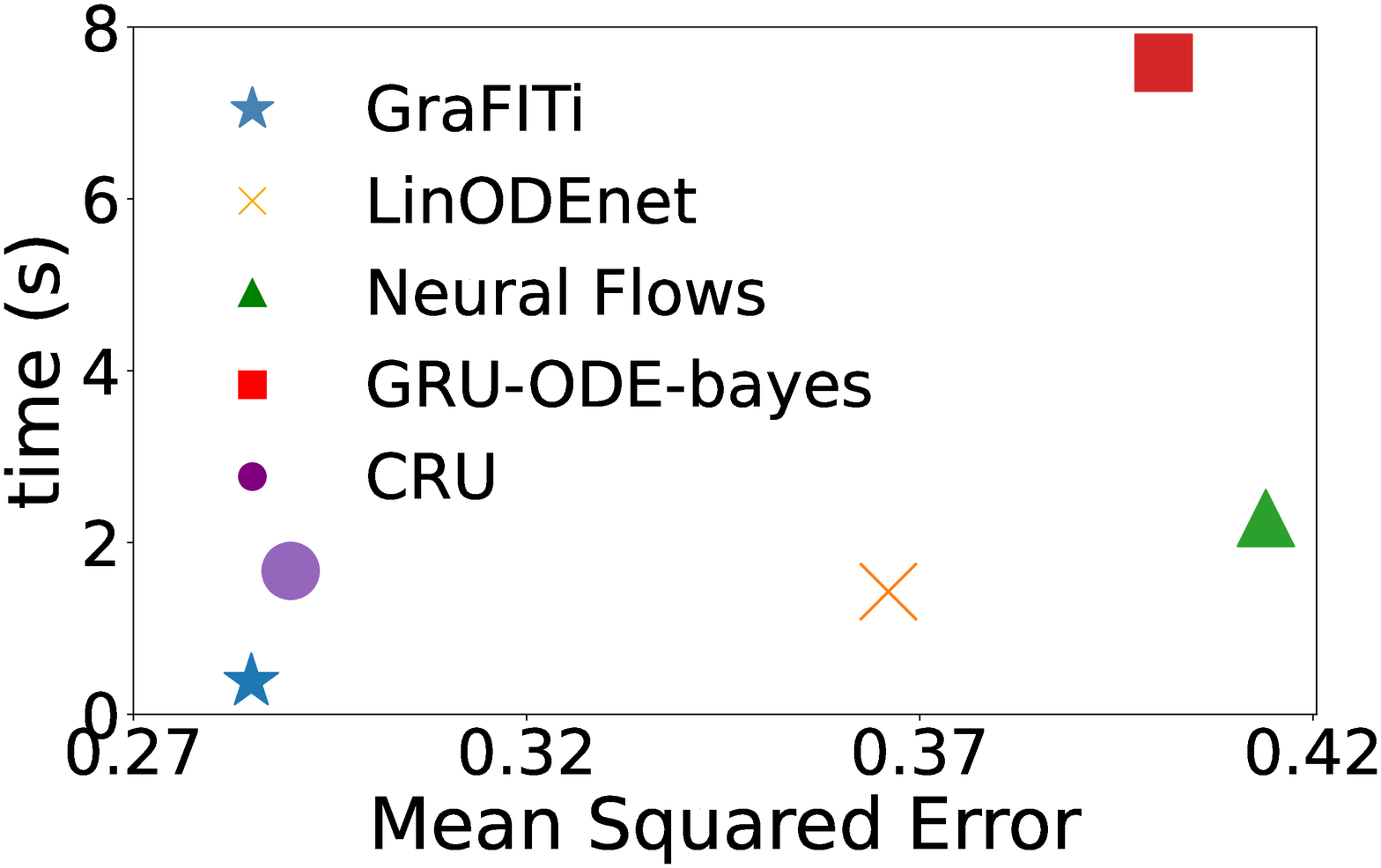}	
	}	
	\subfigure[MIMIC-III]{
		\includegraphics[width=0.45\linewidth]{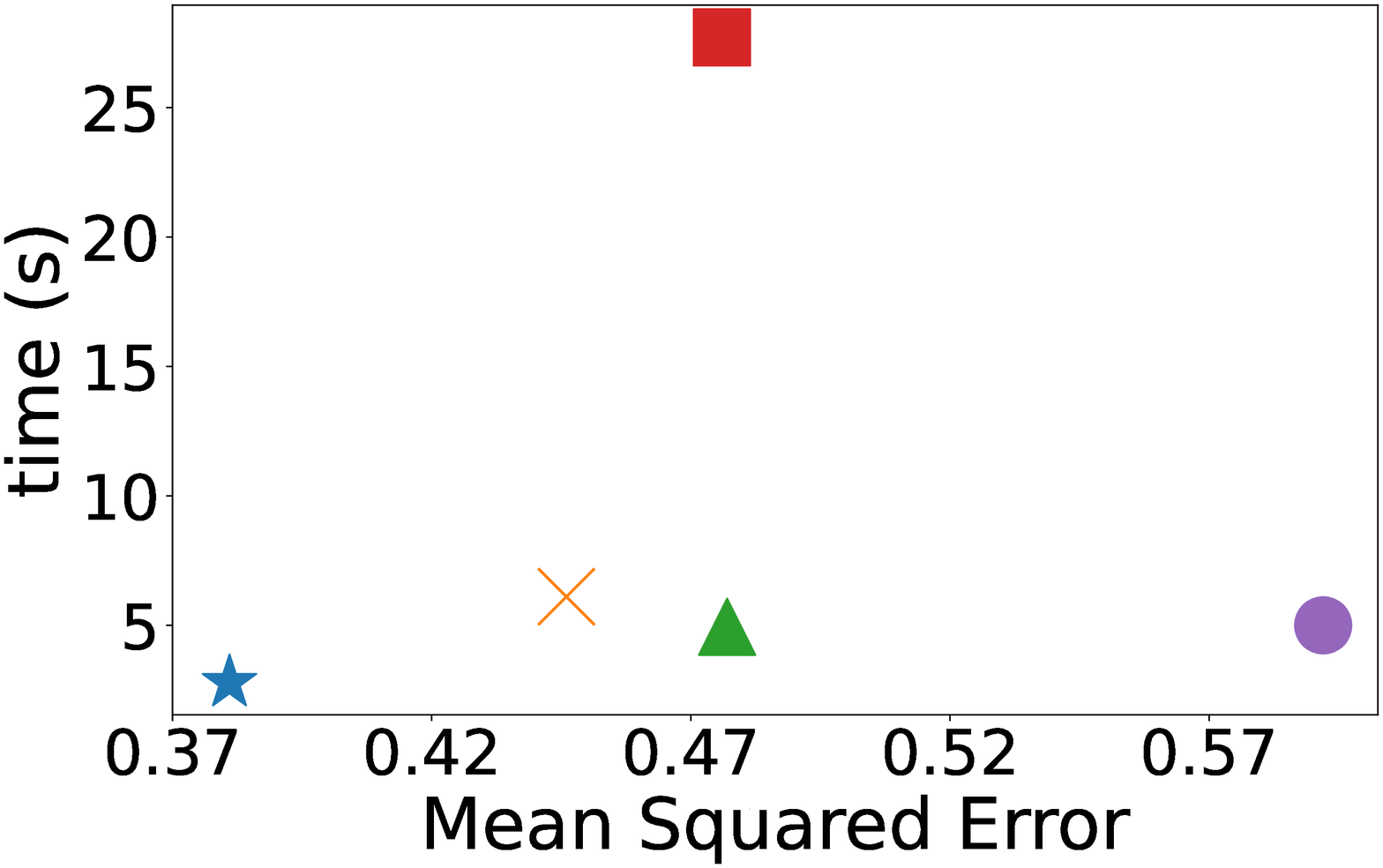}	
	}
	\subfigure[MIMIC-IV]{
		\includegraphics[width=0.45\linewidth]{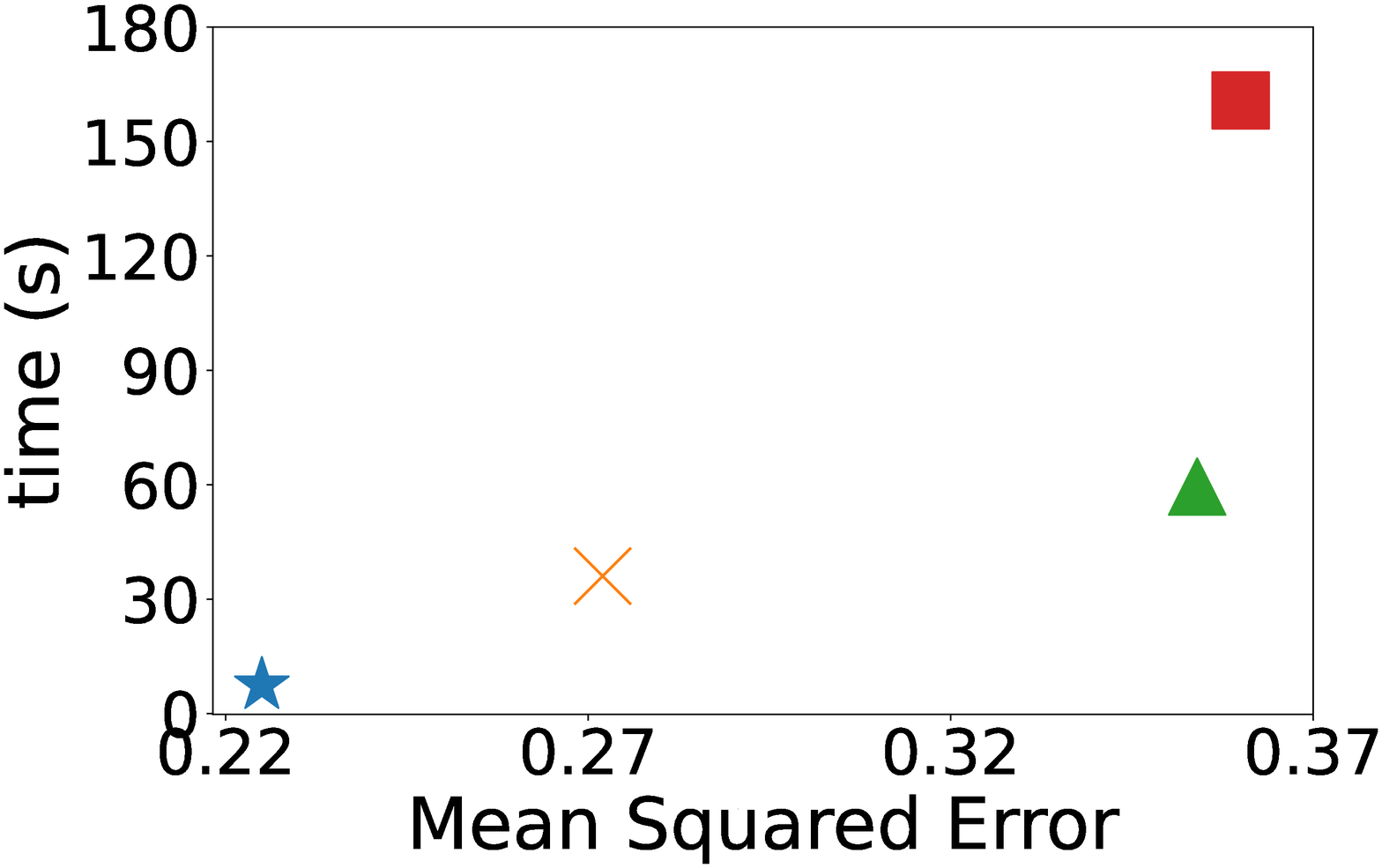}	
	}
	\subfigure[Physionet'12]{
		\includegraphics[width=0.45\linewidth]{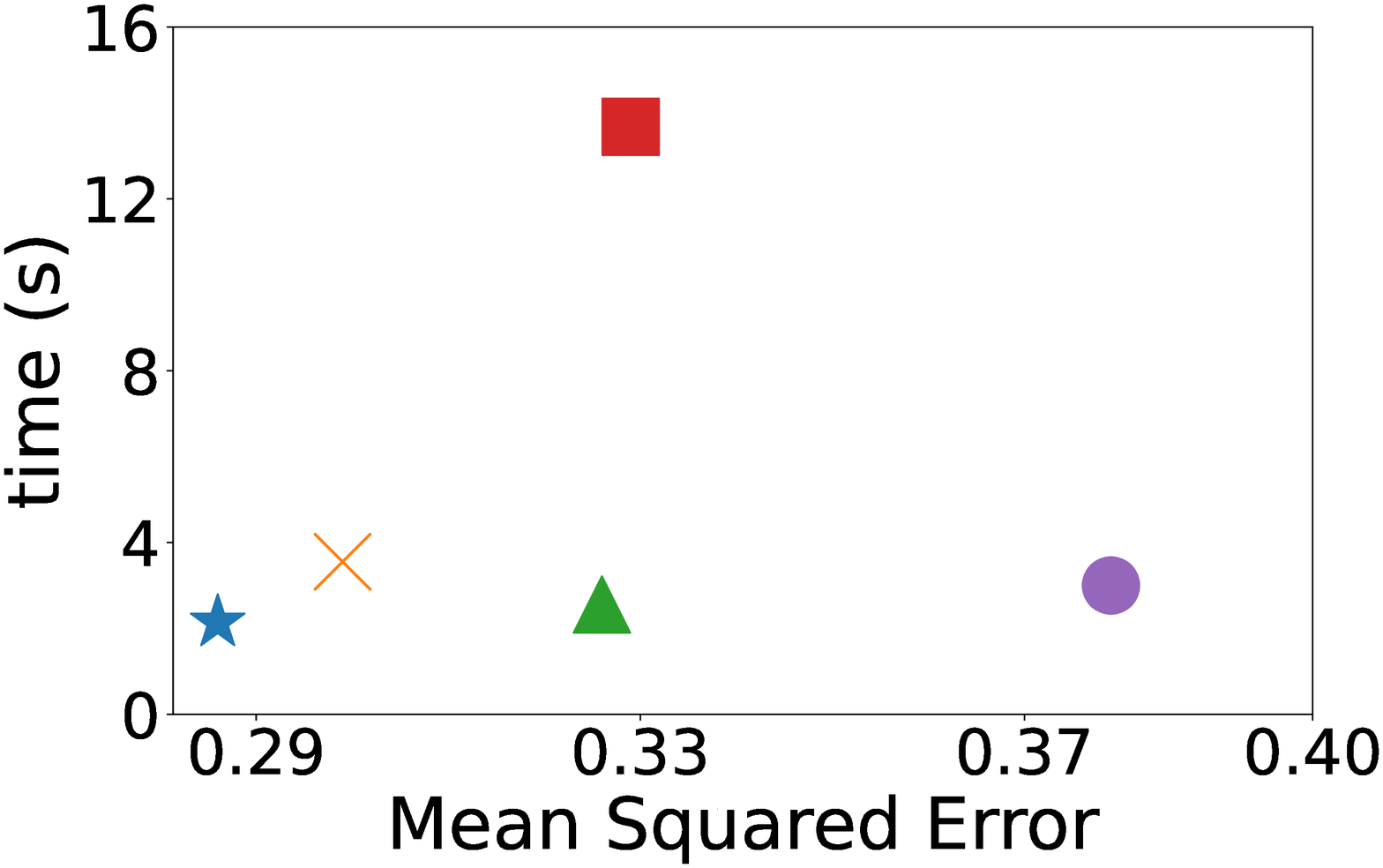}	
	}
	\caption{Comparison of IMTS forecasting models: GraFITi, LinODEnet, CRU, Neural Flows and GRU-ODE-Bayes in terms of efficiency: evaluation time against error metric.}
	\label{fig:efficiency}
\end{figure}

\begin{table*}
	\centering
	\scriptsize
	\caption{Experimental results on varying observation and forecasting ranges for the medical datasets. Evaluation measure is MSE. Lower is better. Best results are in bold and the second best are in italics. ME indicates memory error.}
	\label{tab:var_obs_fore}
	\begin{tabular}{lcc:ccccl}
		&&&&&&\\
		\hline
		&Obs. $/$ Pred.&GraFITi (ours) &LinODEnet	& CRU &\multicolumn{1}{c}{Neural Flow}&\multicolumn{1}{c}{GRU-ODE-Bayes} & $ \quad \uparrow \%$\\
		\hline
		\multirow{4}{*}{MIMIC-III} &	24/12	&$ \mathbf{0.438 \pm 0.009}$	&	{$ \mathit{0.477 \pm 0.021}$}	&	$0.575\pm 0.020$	&	$0.588 \pm 0.014$	&$0.591 \pm 0.018$ &	$ \quad \uparrow \textbf{8.2\%}$ \\
		&24/24	&$ \mathbf{0.491 \pm 0.014} $	&	$\mathit{0.531 \pm 0.022}$&	$0.619\pm0.028$	&	$0.651 \pm 0.017$	&$0.653 \pm 0.023$& 	$\quad \uparrow \textbf{7.5\%}$ \\
		&36/6	&$\mathbf{0.457 \pm 0.050}$		&$\mathit{0.492 \pm 0.019}$	&	$0.647\pm 0.051$	&$0.573 \pm 0.043$	&$0.580 \pm 0.049$ 	&$ \quad \uparrow \textbf{7.1\%}$\\
		&36/12	&$ \mathbf{0.490 \pm 0.027}$	&	$\mathit{0.554 \pm 0.042}$&	$0.680\pm0.043$	&	$0.620 \pm 0.035$&	$0.632 \pm 0.044$ &	$\quad \uparrow \textbf{10.8\%}$\\
		\hline							
		\multirow{4}{*}{MIMIC-IV}	&24/12&	$ \mathbf{0.285 \pm 0.001}$	&	$\mathit{0.335 \pm 0.002}$	&	ME	&$0.465 \pm 0.003$	&$0.366 \pm 0.154$ &	$ \quad \uparrow \textbf{14.9\%}$\\
		&24/24	&$ \mathbf{0.285 \pm 0.002}  $	&	$\mathit{0.336 \pm 0.002}$&	ME	&	$0.465 \pm 0.003$&	$0.439 \pm 0.003$ &	$\quad \uparrow \textbf{15.1\%}$\\
		&36/6	&$ \mathbf{0.260 \pm 0.002} $	&	$\mathit{0.309 \pm 0.002}$&	ME	&	$0.405 \pm 0.001$&	$0.393 \pm 0.002$ &	$\quad \uparrow \textbf{15.9\%}$\\
		&36/12	&$ \mathbf{0.261 \pm 0.005}$	&	$\mathit{0.309 \pm 0.002}$&	ME	&	$0.395 \pm 0.001$&	$0.393 \pm 0.002$ &	$ \quad \uparrow \textbf{15.5\%}$\\
		\hline							
		\multirow{4}{*}{Physionet'12}	&24/12	&$ \mathbf{0.365 \pm 0.001}$&	$\mathit{0.373 \pm 0.001}$	&	$0.435\pm0.001$	&$0.431 \pm 0.001$&	$0.432 \pm 0.003$ &	$ \quad \uparrow \textbf{2.1\%}$ \\
		&24/24	&$ \mathbf{0.401 \pm 0.001}$	&	$\mathit{0.411 \pm 0.001}$	&	$0.467\pm0.002$	&$0.506 \pm 0.002$	&$0.505 \pm 0.001$ &	$ \quad \uparrow \textbf{2.4\%}$ \\
		&36/6	&$ \mathbf{0.319 \pm 0.001} $	&	$\mathit{0.329 \pm 0.001}$	&	$0.396\pm0.003$	&$0.365 \pm 0.001$	&$0.363 \pm 0.004$ &	$\quad \uparrow \textbf{3.0\%}$ \\
		&36/12	&$ \mathbf{0.347 \pm 0.001} $	&	$\mathit{0.357 \pm 0.001}$	&	$0.417\pm0.001$	&$0.398 \pm 0.001$	&$0.401 \pm 0.003$ &	$\quad \uparrow \textbf{2.8\%}$ \\

		\hline
		& & & & && &\\
	\end{tabular}
\end{table*}

\paragraph{Efficiency comparison}

We compare the efficiency of leading IMTS forecasting models: GraFITi, LinODEnet, CRU, Neural Flow, and GRU-ODE-Bayes. We evaluate them in terms of both execution time (batch size: 64) and MSE. The results, presented in Figure~\ref{fig:efficiency}, show that for datasets with longer time series like MIMIC-IV and USHCN, GraFITi significantly outpaces ODE and flow-based models. Specifically, GraFITi is over $5$ times faster than the fastest ODE model, LinODEnet, in both MIMIC-IV and USHCN. Even for shorter time series datasets like Physionet'12 and MIMIC-III, GraFITi remains twice as fast as LinODEnet. Moreover, on average, GraFITi surpasses GRU-ODE-Bayes in speed by an order of magnitude.

\paragraph{Varying observation and forecast ranges}

This experiment is conducted with two different observation ranges ($24$ and $36$ hours) and two different prediction ranges for each observation range. Specifically, for the observation range of $24$ hours, the prediction ranges are $12$ and $24$ hours, and for the observation range of $36$ hours, the prediction ranges are $6$ and $12$ hours. This approach allows for a more comprehensive evaluation of the model's performance across various scenarios of observation and prediction ranges. The results are presented in Table~\ref{tab:var_obs_fore}.

Again for varying observation and forecast ranges, GraFITi is the top performer, followed by LinODENet. Significant gains in forecasting accuracy are observed in the MIMIC-III and MIMIC-IV datasets. {\em On average, GraFITi improves the accuracy of LinODEnet, the next best IMTS forecasting model, by $8.5\%$ in MIMIC-III, $15.5\%$ in MIMIC-IV, and $2.6\%$ in the Physionet'12 dataset}.


\begin{table}[ht]
	\centering
	\scriptsize
	\caption{Performance of GraFITi with varying sparsity levels using MIMIC-III dataset. The `IMTS' dataset refers to the actual dataset, while `AsTS' is a synthetic asynchronous time series dataset created by restricting the number of observed channels at each time point to $1$. The `AsTS + x\%' dataset is created by modifying `AsTS' dataset by retrieving x\% of the missing observations. Goal is to observe $36$ hours of data and then forecast the next $3$ time steps}
	\label{tab:var_spars}
	\begin{tabular}{lccccc}
		\hline
		Model	& IMTS	& AsTS	&	AsTS+10\%	& AsTS+50\%	&	AsTS+90\% \\
		\hline
		GraFITi	&	{$\mathbf{0.396}$}	& $0.931$	& $0.845$ & $\mathbf{0.547}$ & {$\mathbf{0.413}$} \\
				LinODENet	& $0.446$	& {$\mathbf{0.894}$}	& {$\mathbf{0.815}$} & $0.581$ & $0.452$\\
		\hline
	\end{tabular}
\end{table}

\subsection{Limitations}
\label{sec:discussion}

The GraFITi model is a potential tool for forecasting on IMTS.
It outperforms existing state-of-the-art models, even in highly sparse datasets (up to $98\%$ sparsity in MIMIC-IV).
However, we note a limitation in the compatibility of the model with certain data configurations.
In particular, the GraFITi faces a challenge when applied to Asynchronous Time Series datasets.
In such datasets, channels are observed asynchronously at various time points, resulting in disconnected sparse graphs.
This disconnection hinders the flow of information and can be problematic
when channels have a strong correlation towards the forecasts as model may not be able to capture these correlations.
It can be observed from Table~\ref{tab:var_spars}
where GraFITi is compared with the next best baseline model LinODENet for varying sparsity levels using MIMIC-III dataset.
The performance of GraFITi deteriorates with increase in sparsity levels
and gets worst when the series become asynchronous.

Moreover, the existing model cannot handle meta data associated with the IMTS.
Although, these meta data points could be introduced as additional channel nodes,
this would again disconnect the graph due to lack of edges connecting the time nodes to meta data nodes.
One possible solution to both the challenges is
to interconnect all the channel nodes including meta data if present,
and apply a distinct multi-head attention on them.
Therefore, in future, we aim to enhance the capability of the proposed model
to handle Asynchronous Time Series datasets and meta data in the series.

%% file: section-7-conclusions.tex
\section{Conclusions}

In this paper, we propose a Graph based model called GraFITi
for the forecasting of irregularly sampled time series with missing values (IMTS).
First we represent the time series as a Sparsity Structure Graph
with channels and observation times as nodes and observation measurements as edges;
and re-represent the task of time series forecasting as an edge weight prediction problem in a graph.
An attention based architecture is proposed
for learning the interactions between the nodes and edges in the graph. 
We experimented on $4$ datasets including $3$ real world and $1$ synthetic dataset
for various observation and prediction ranges.
The extensive experimental evaluation demonstrates that
the proposed GraFITi provides superior forecasts compared to the
state-of-the-art IMTS forecasting models.

%% file: apdx.tex
\newpage

\appendix

\section{Ablation studies}

\subsection{Importance of target edge}

In the current graph representation,
target edges are connected in the graph providing rich embedding to learn the edge weight.
However, we would like to see the performance of the model without query edges in the graph.
For this, we compared the experimental results of GraFITi and GraFITi$\setminus$T
where GraFITi$\setminus$T is the same architecture without query edges in the graph.
The predictions are made after $L-1^{\text{th}}$ layer
by concatenating the channel embedding and the sinusoidal embedding
of the query time node and passing it through the dense layer.
The results for the real IMTS datasets are reported in Table~\ref{tab:grafiti-t}.

\begin{table}[h]
	\centering
	\scriptsize
	\caption{Performance of GraFITi$\setminus$T (GraFITi without query edges in the graph), evaluation metric MSE.}
	\label{tab:grafiti-t}
	\begin{tabular}{lccc}
		\toprule
		Model & MIMIC-III & MIMIC-IV & Physionet'12 \\
		GraFITi & {$\mathbf{0.396 \pm 0.030}$} & {$\mathbf{0.225 \pm 0.001}$}& {$\mathbf{0.286 \pm 0.001}$}\\
		GraFITi$\setminus$T & $0.433 \pm 0.019$& $0.269 \pm 0.001$& $0.288 \pm 0.001$ \\
		\hline
	\end{tabular}
\end{table}

We see that the performance of the GraFITi deteriorated significantly without query edges in the graph.
GraFITi exploit the sparse structure in the graph for predicting the weight of the query edge.
The attention mechanism help the target edge to gather the useful information from the incident nodes.

\subsection{MAB vs GAT as Graph Neural Network module {\gnn} (Eq~\ref{eq:gnn(l)})}

\begin{table}[h]
	\centering
	\scriptsize
	\caption{Comparing performance of MAB (GraFITi-MAB) and GAT (GraFITi-GAT) as \gnn{} module in GraFITi.}
	\label{tab:mab_gat}
	\begin{tabular}{lccc}
					&	MIMIC-III			&	MIMIC-IV			&	Physionet'12	\\
					\hline
		GraFITi-MAB &	$0.396 \pm 0.030$	&	$0.225 \pm 0.001$	&	$0.286 \pm 0.001$\\
		GraFITi-GAT	&	$0.388 \pm 0.020$	&	$0.225 \pm 0.001$	&	$0.288 \pm 0.001$\\
	\end{tabular}
\end{table}

In Table~\ref{tab:mab_gat}, we compare the performance of MAB and GAT as a \gnn{} module in the proposed GraFITi.
We use MIMIC-III, MIMIC-IV and Physionet'12 for the comparison.
We notice that the performance of MAB and GAT are similar. Hence, we use MAB as the \gnn{} module in the proposed GraFITi.
{\em Please note that the objective of this work is not to find the best \gnn{} module
but to show IMTS forecasting using graph neural networks.}

\section{Hyperparameter search}
\label{sec:hp_search}

We search the following hyperparameters for IMTS forecasting models as mentioned in the respective works:

\paragraph{GRU-ODE-Bayes:}
We set the number of hidden layers to $3$ and selected solver from $\{\text{euler}, \text{dopri}\}$

\paragraph{Neural Flows}
We searched for the flow layers from $\{1,4\}$ and set the hidden layers to $2$

\paragraph{LinODENet}
We searched for hidden size from $\{64,128\}$, latent size from $\{128,192\}$. We set the encoder with $5$-block ResNet with 2 ReLU pre-activated layers each, StackedFilter of 3 KalmanCells, with linear one in the beginning.

\paragraph{CRU}
We searched for latent state dimension from $\{10,20,30\}$, number of basis matrices from $\{10, 20\}$ and bandwidth from $\{3,10\}$.

\paragraph{mTAN}
We searched the \#attention heads from $\{1,2,4\}$, \#reference time points from $\{8,16,32,64,128\}$, latent dimensions form $\{20,30,40,50\}$, generator layers from $\{25,50,100,150\}$, and reconstruction layers from $\{32,64,128,256\}$.

%

For the all the MTS forecasting models, we used the default hyperparameters provided in~\cite{ZC22}.